\PassOptionsToPackage{dvipsnames}{xcolor}

\documentclass{article}
\usepackage[utf8]{inputenc} 
\usepackage[T1]{fontenc}    
\usepackage{hyperref}       
\usepackage{url}            
\usepackage{booktabs}       
\usepackage{amsfonts}       
\usepackage{nicefrac}       
\usepackage{microtype}      
\usepackage{graphicx}
\usepackage{float}
\usepackage{natbib}
\usepackage{doi}
\usepackage{xcolor}
\usepackage{lineno}
\usepackage{amsmath}
\usepackage{xcolor}
\usepackage{enumitem}
\usepackage{arxiv}
\usepackage{url}
\newcommand\myshade{85}
\colorlet{mylinkcolor}{RoyalBlue}
\colorlet{mycitecolor}{violet}
\colorlet{myurlcolor}{YellowOrange}

\hypersetup{
  linkcolor  = mylinkcolor!\myshade!black,
  citecolor  = mycitecolor!\myshade!black,
  urlcolor   = myurlcolor!\myshade!black,
  colorlinks = true,
}

\usepackage{color}

\usepackage{times}
\usepackage{latexsym}

\usepackage[T1]{fontenc}

\usepackage[utf8]{inputenc}

\usepackage{microtype}

\usepackage{booktabs}
\usepackage{graphicx}
\usepackage{makecell}
\usepackage{multirow}
\usepackage{longtable}
\usepackage{CJKutf8}
\usepackage{float}
\usepackage{algorithm}
\usepackage{algpseudocode}
\usepackage[frozencache, cachedir=minted-cache]{minted}

\usepackage{tikz}
\usetikzlibrary{shapes.geometric}

\title{
ChatDB: Augmenting LLMs with Databases as Their Symbolic Memory
}

\newcommand*\samethanks[1][\value{footnote}]{\footnotemark[#1]}

\author{
\small
    Chenxu Hu\textsuperscript{1}\begin{NoHyper}\thanks{Equal technical contribution.}\end{NoHyper}\quad 
    Jie Fu\textsuperscript{2}\samethanks[1]~~\begin{NoHyper}\thanks{Corresponding authors.}\end{NoHyper}\quad 
    Chenzhuang Du\textsuperscript{1}\quad 
    Simian Luo\textsuperscript{1}\quad 
    Junbo Zhao\textsuperscript{3}\quad
    Hang Zhao\textsuperscript{1}\samethanks[2]
\\
\small
    \textsuperscript{1}Tsinghua University\quad \textsuperscript{2}Beijing Academy of Artificial Intelligence \quad 
    \textsuperscript{3}Zhejiang University
\\
\small
\texttt{fujie@baai.ac.cn} \quad 
\texttt{hangzhao@mail.tsinghua.edu.cn} 
}

\begin{document}

\maketitle

\begin{abstract}


Large language models (LLMs) with memory are computationally universal~\citep{schuurmans2023memory}. 
However, mainstream LLMs are not taking full advantage of memory, and the designs are heavily influenced by biological brains. 
Due to their approximate nature and proneness to the accumulation of errors, conventional neural memory mechanisms cannot support LLMs to simulate complex reasoning.  
In this paper, we seek inspiration from modern computer architectures to augment LLMs with symbolic memory for complex multi-hop reasoning. 
Such a symbolic memory framework is instantiated as an LLM and a set of SQL databases, where the LLM generates SQL instructions to manipulate the SQL databases. 
We validate the effectiveness of the proposed memory framework on a synthetic dataset requiring complex reasoning. 
The project website is available at \url{https://chatdatabase.github.io/}.

\end{abstract}

\begin{figure*}[h]
\begin{centering}
\includegraphics[width=\textwidth]{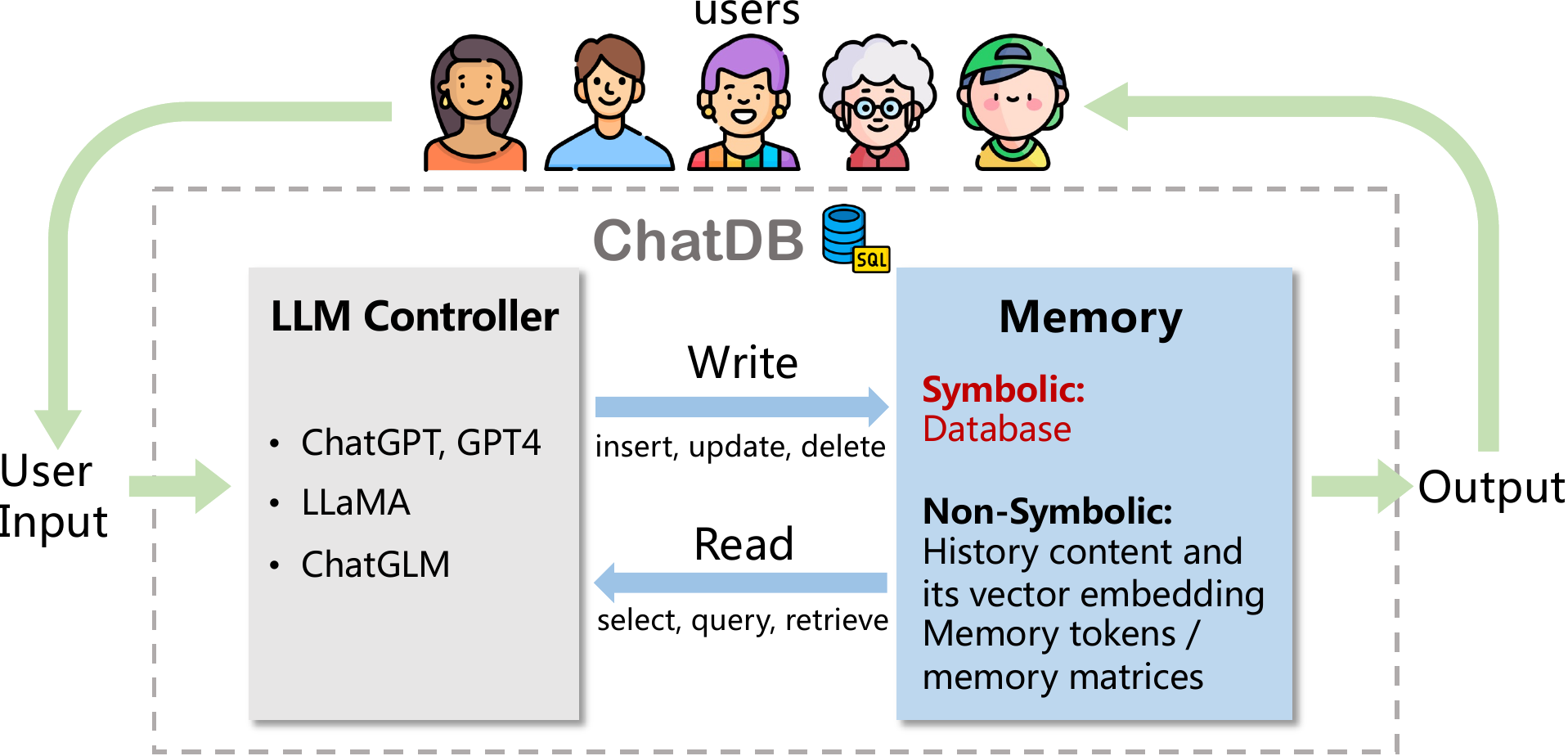}
\caption{Overall workflow of ChatDB. 
The LLM controller controls the read and write operations to the memory. The memory stores historical information and provides relevant historical information to assist in responding to user input. In ChatDB, we focus on augmenting LLMs with databases as their symbolic memory.}\label{fig:teaser}
\end{centering}
\end{figure*}

\section{Introduction}
Large language models (LLMs), such as GPT-4~\citep{openai2023gpt4} and PaLM 2~\citep{anil2023palm}, have increasingly become an essential component of modern artificial intelligence (AI) systems, revolutionizing our understanding of natural language processing (NLP) and transforming various industries~\citep{hao2023reasoning,wang2023visionllm}. 
While LLMs have made significant strides in understanding and generating contextually relevant responses, they have limitations~\citep{chen2023two}. 
One of the main challenges is that multi-turn interactions with language models generate a large number of tokens, which can easily exceed the input token limit of LLMs. 
For example, GPT-4 (32K) can only handle 32,000 tokens.
As the interaction progresses, the LLMs must maintain contextual information (\textit{e.g.}, user inputs, and previous responses) and generate responses based on the accumulated data. 
Simply concatenating all contextual information and cramming it into LLMs, however, can easily exceed the processing capabilities of LLMs and accumulate errors, causing the model to lose track of the conversation and produce less accurate responses.

Some neural memory mechanisms have been explored~\citep{wu2022memorizing,khattab2022demonstrate,zhong2022training} to overcome the limited token input issue of LLMs. 
The memory components serve as a storage and retrieval system for relevant information from previous interactions. 
However, augmenting LLMs with conventional neural memory usually leads to difficulties in storing, retrieving, and manipulating historical information in memory, especially for tasks requiring complex multi-hop reasoning. 
Two main causes are (a) They do not store historical information in a structured form; (b) Their manipulation of the information stored in memory is not symbolic, as they all rely on some vector similarity calculations, which can be inaccurate, thus leading to the accumulation of errors. 

To address the aforementioned issues, we propose using databases as novel symbolic memory for LLMs. 
The whole framework is named \textbf{ChatDB}. 
As shown in Figure~\ref{fig:teaser}, ChatDB consists of two components: an LLM controller and its memory. 
The LLM controller can be any commonly used LLM~\citep{openai2023gpt4,touvron2023llama,du2022glm,zeng2022glm} and is responsible for controlling the read and write operations to the memory. 
The memory of LLMs, which can be symbolic or non-symbolic, or a combination of both, is responsible for storing historical information and providing information when needed to assist the LLM in responding to user input.
In ChatDB, we focus on using databases as symbolic memory, which allows for the structured storage of historical information through the execution of a symbolic language, namely SQL statements. 
These SQL statements are generated by an LLM. 
Incorporating a database as symbolic memory is particularly useful in scenarios requiring precise recording, modification, querying, deletion, and analysis of historical data. 
For example, a store manager needs to maintain daily sales records, where using plain text or matrices as memory is unsuitable~\citep{chen2023two}. However, using a database as external symbolic memory is highly suitable.
The database enables accurate operations, including data insertion, deletion, update, and selection, using SQL statements. 
Thus, employing databases as external symbolic memory ensures precision and efficiency in managing and manipulating historical data, significantly enhancing the performance of LLMs in scenarios that require high accuracy and long-term data recording and processing.

In the ChatDB framework, we propose the \textit{chain-of-memory}~(CoM) approach to manipulate the external symbolic memory more effectively, thereby further enhancing the reasoning capabilities of LLMs.
The chain-of-memory approach transforms user input into a series of intermediate memory operation steps that lead to final results. Through the chain-of-memory approach, a complex problem is decomposed into multiple steps of memory operations, significantly reducing the complexity of problem-solving. In ChatDB, each intermediate step involves one or more SQL statements.

Our ChatDB makes several contributions to the field of LLMs. 
Firstly, we propose augmenting LLMs with databases as their external symbolic memory, allowing for structured storage of historical data and enabling symbolic and complex data operations using SQL statements.
Secondly, our chain-of-memory approach enables effective memory manipulation by converting user input into multi-step intermediate memory operations, which enhance the performance of ChatDB, enabling it to handle complex, multi-table database interactions with improved accuracy and stability.
Finally, our experiments demonstrate that augmenting LLMs with symbolic memory improves multi-hop reasoning capabilities and prevents error accumulation, thereby enabling ChatDB to significantly outperform ChatGPT on a synthetic dataset.

\section{Related Work}
\textbf{Memory-Augmented Large Language Models~(LLMs).}
LLMs, such as GPT-4~\citep{openai2023gpt4} and PaLM 2~\citep{anil2023palm}, have demonstrated powerful reasoning and decision-making abilities. 
However, LLMs are often hindered by their limited context window sizes (\textit{e.g.}, GPT-4 can only handle 32K tokens).
Memory-augmented LLMs~\citep{wu2022memorizing,wu2022efficient,zhong2022training,lewis2020retrieval,guu2020retrieval,park2023generative,khattab2022demonstrate,izacard2022few} incorporate a memory module that prevents the model from forgetting crucial information and allows it to handle long text inputs that exceed the context window size. 
Retrieval-augmented in-context learning~\citep{khattab2022demonstrate} uses retrieval models (RM) to retrieve
relevant information that can be inserted into the LLM as a prompt.
For example, Auto-GPT~\footnote{\url{https://github.com/Significant-Gravitas/Auto-GPT}} and Generative Agents~\citep{park2023generative} utilize a memory module to store the text prompt directly, allowing the agent to keep track of its history. The past and current prompts are then input into the LLM for processing.
Neural Turing Machines (NMT)~\citep{graves2014neural}, which incorporate the recurrent neural network (RNN) with external trainable memory resources and learn to interact with the memory module with gradient descent. 
Gated Graph Sequence Neural Network~(GGS-NN)~\citep{johnson2017learning} constructs and modifies graphs and utilizes the graphs to produce reasonable outputs.
Recurrent Memory Transformer (RMT)~\citep{bulatov2022recurrent} introduces additional memory tokens to the input and output sequences to store, process and exchange local and global information between segments of long sequences, and then train the model to control both memory operation and sequence representations processing. 


\textbf{Reasoning with LLMs.}
LLMs are known to struggle in complex reasoning tasks. 
Previous methods focus on incorporating specially designed supervisory signals or fine-tuning to enhance the reasoning ability of language models~\citep{pikekos2021measuring,ran2019numnet,andor2019giving,cobbe2021training,chen2022program}. 
Recent methods mainly improve the reasoning ability of language models through In-Context Learning~\citep{brown2020language, lester2021power,wei2021finetuned,wei2022chain,wang2022self}.
The most representative of these is Chain-of-Thought~(CoT)~\citep{wei2022chain}, which presents the intermediate reasoning process of solving sample problems to the language model, greatly enhancing its reasoning capabilities.

\textbf{LLMs with DBs.} LLMs have demonstrated an impressive capability in generating code, including Python code, execution commands for Excel, and Structured Query Language (SQL) for databases~\citep{openai2023gpt4}. 
ChatExcel~\footnote{\url{https://chatexcel.com/}} uses LLMs to generate the Excel execution command, simplifying the user interaction process. 
BINDER~\citep{cheng2022binding} proposes a framework that maps task inputs to executable programs in a programming language (\textit{e.g.}, Python code) bound with an API to call LLMs to perform a wide range of functionalities. 
SQL-PALM~\citep{sun2023sqlpalm} proposes an LLM-based Text-to-SQL model, using the execution-based self-consistent prompting approach, and outperforms previous Text-2-SQL methods by a large margin. 
While previous works involve databases to some extent, our proposed ChatDB system differs significantly from these methods. 
In specific, ChatDB views the databases as the external symbolic memory module for the LLM, and then leverages the database for reading and writing essential data information to enhance the reasoning process via \textit{chain-of-memory}, leading to more accurate reasoning results.

\textbf{Tool-using LLMs.}
From the tool-using perspective, ChatDB can also be seen as an LLM utilizing DBs as a tool~\citep{schick2023toolformer,shen2023hugginggpt,suris2023vipergpt,paranjape2023art}.
Toolformer~\citep{schick2023toolformer}, through a series of demonstrations, instructs the language model that it can invoke some APIs to utilize external tools to solve the current problem.
Another representative work is Auto-GPT~\footnote{https://github.com/Significant-Gravitas/Auto-GPT}, which enables the language models to complete a series of impressive tasks using a search engine.
The advantage of ChatDB, which uses a database as an external tool, is that it allows the language model to maintain more accurate records and use historical data, thereby addressing more complex problems, especially those that require accurate historical data for reasoning.

\section{ChatDB}
In this section, we first briefly introduce the task definition and setting. We then describe the overall framework of our proposed ChatDB. Finally, we delve into the details of the chain-of-memory approach, which is the main component of ChatDB.

\subsection{Task Definition}
Given a user input in natural language and details of existing tables in the database (not required if there are no existing tables), the goal is to manipulate the symbolic memory, \textit{i.e.}, the external database, to fulfill the user's request. 
For instance, if the user (\textit{e.g.}, a store manager) command is to record, modify, query, and delete specific data, the corresponding SQL operation should be to insert, update, select, and delete the relevant data in the appropriate tables, respectively. These operations often involve multiple tables in the database.

\begin{figure*}[t]
\begin{centering}
\includegraphics[width=\textwidth]{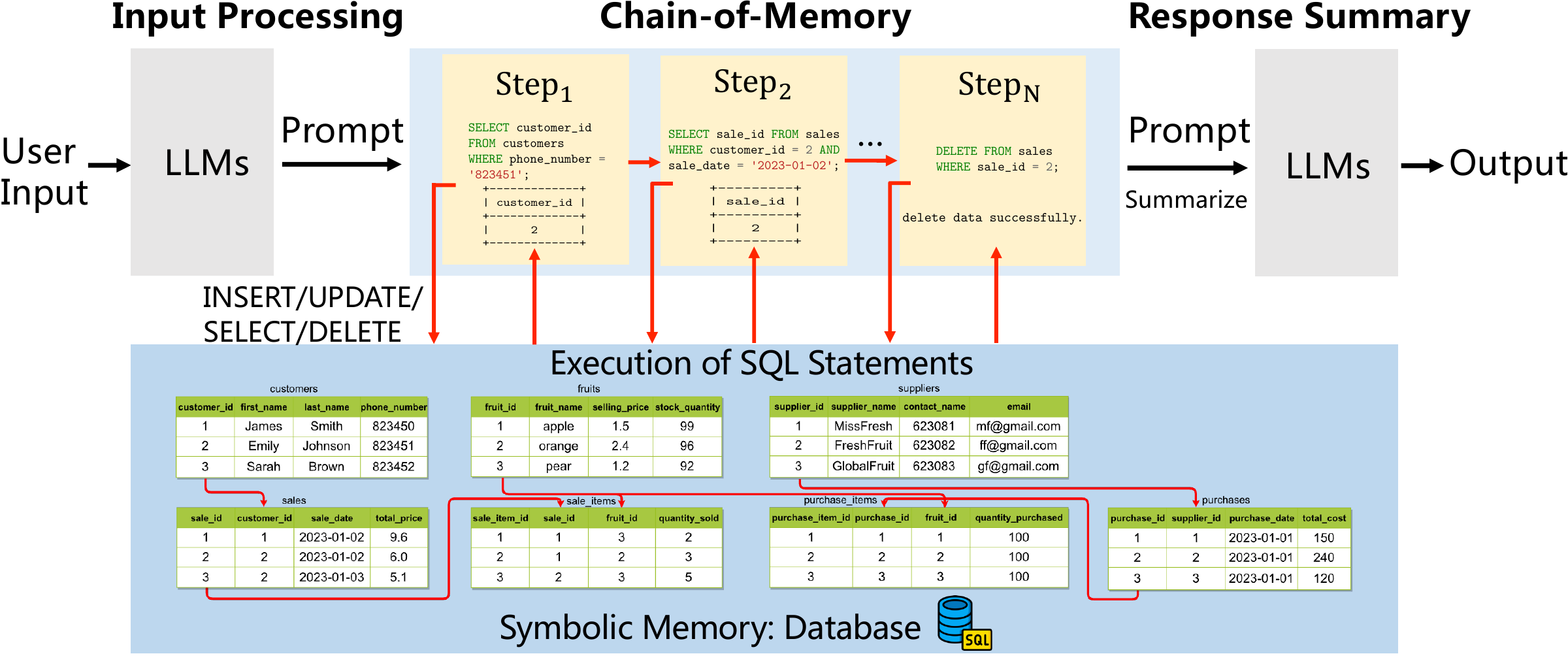}
\caption{ChatDB framework. The red arrow lines represent the process flow of chain-of-memory, indicating the connection between multiple memory operations. The red arrow lines between database tables represent the reference relationships between primary keys and foreign keys, which start from primary keys to foreign keys. Only the first four columns of each table are shown for brevity. This example showcases the process of returning goods purchased on 2023-01-02 by a customer with the phone number 823451. \label{fig:chatdb}}
\end{centering}
\end{figure*}

\subsection{Framework Overview}
ChatDB framework consists of three main stages: input processing, chain-of-memory, and response summary, as illustrated in Figure~\ref{fig:chatdb}.
Algorithm~\ref{alg:chatdb} provides a detailed illustration of the entire algorithmic procedure of ChatDB to respond to the user input.

\textbf{Input Processing.} ChatDB generates a series of intermediate steps to manipulate the symbolic memory by utilizing LLMs if responding to the user input requires the use of memory. Otherwise, we use LLMs directly to generate a reply.

\textbf{Chain-of-Memory.} ChatDB executes a series of intermediate memory operation steps to interact with symbolic memory. ChatDB manipulates the symbolic memory in sequence according to a series of previously generated SQL statements, including operations such as insert, update, select, delete, etc. The external database executes the corresponding SQL statements, updates the database, and returns the results. It is worth noting that ChatDB decides whether to update the memory operation step based on the results of previous SQL statements before executing this operation. ChatDB executes the next step following the same procedure until all operations on the memory are completed.

\textbf{Response Summary.} ChatDB summarizes the final response to the user based on the results of a series of chain-of-memory steps.

\begin{algorithm}
\caption{The algorithm of ChatDB}\label{alg:chatdb}
\begin{algorithmic}[1]
\Require \texttt{userInput}, \texttt{dataBase} 
\Ensure \texttt{reply}

\Comment{Begin of Input Processing}

\If{need manipulate memory to respond to \texttt{userInput}}
    \State \texttt{memOps} = \textcolor[HTML]{3078BE}{$\mathrm{LLM_{getSteps}}$}(\texttt{userInput})  \Comment{Generate intermediate steps using LLMs}
\Else
    \State \texttt{reply} = \textcolor[HTML]{3078BE}{$\mathrm{LLM}$}(\texttt{userInput})  \Comment{Generate reply using LLMs directly}
    \State \Return \texttt{reply}
\EndIf

\Comment{Begin of Chain-of-Memory}

\State \texttt{sqlResults} = \texttt{[]}
\State \texttt{newMemOps} = \texttt{[]}
\For{each \texttt{memOp} in \texttt{memOps}}  

    \If{need update \texttt{memOp} based on \texttt{sqlResults}} 
        \State \texttt{newMemOp} = \textcolor[HTML]{3078BE}{$\mathrm{LLM_{updateOperation}}$}(\texttt{memOp}, \texttt{sqlResults})
    \Else
        \State \texttt{newMemOp} = \texttt{memOp}
    \EndIf
    \State \texttt{sqlResult} = \textcolor[HTML]{3078BE}{$\mathrm{executeOperation}$}(\texttt{newMemOp}, \texttt{dataBase})  \Comment{Execute operation on database}
    \State \texttt{sqlResults}.\textcolor[HTML]{3078BE}{$\mathrm{append}$}(\texttt{sqlResult})       
    \State \texttt{newMemOps}.\textcolor[HTML]{3078BE}{$\mathrm{append}$}(\texttt{newMemOp})
\EndFor

\Comment{Begin of Response Summary}

\State \texttt{reply} = \textcolor[HTML]{3078BE}{$\mathrm{LLM_{summary}}$}(\texttt{userInput}, \texttt{newMemOps}, \texttt{sqlResults})  \Comment{Summarize the final reply}
\State \Return \texttt{reply}

\end{algorithmic}
\end{algorithm}

\subsection{Chain-of-Memory}

Chain-of-thought~\citep{wei2022chain} emphasizes decomposing complex reasoning into a series of intermediate steps. 
Chain-of-memory~(CoM) can be seen as a way to augment chain-of-thought by providing a symbolic memory mechanism to support the storage associated with those intermediate steps. 

The purpose of chain-of-memory is to enhance the reasoning capabilities and robustness of LLMs when manipulating symbolic memory. The approach involves converting the user input into a sequence of intermediate memory operations, enabling LLMs to more accurately and effectively manipulate the memory in a symbolic way. 
The ability to manipulate symbolic memory is particularly valuable for real-world applications that involve complex and accurate interactions with historical data, such as record-keeping and data analysis in a management setting. 

To enhance the performance and stability of our approach, we employ in-context learning~\citep{brown2020language}, providing prompt exemplars of several sequences of chain-of-memory steps, and chain-of-thought prompting.
A robust and accurate chain-of-memory process enables LLMs to better reason over symbolic memory and handle more complex scenarios.

The advantages of chain-of-memory are twofold. Firstly, it enables LLMs to perform complex database manipulations with greater accuracy, enhancing their multi-hop reasoning ability over symbolic memory. Secondly, by decomposing complex operations into a sequence of intermediate memory operations, the chain-of-memory approach enhances the capability of LLMs when handling complex, multi-table interactions. This approach enables LLMs to handle edge cases and unexpected scenarios better, making it a promising method for real-world applications.

\subsection{Comparison with previous memory-augmented LLMs}

\begin{table}[ht]
\scriptsize
\centering
\caption{Comparison with prompt-based memory and matrix-based memory.}
\label{tab:cmp}
\setlength{\tabcolsep}{7pt}{
\begin{tabular}{@{}llllllll@{}}
\toprule
Type     & Model  & Memory format    & Supported operations & Memory storage & Memory execution & Interpretability & State tracking \\ \midrule
Symbolic & 
ChatDB & 
\begin{tabular}[c]{@{}l@{}}Symbolic storage\\ (\textit{e.g.}, database)\end{tabular} & 
\begin{tabular}[c]{@{}l@{}}Insert, Delete\\ Update, Select\end{tabular} &
Structured & 
Symbolic & 
High & 
Yes \\
Prompt-based &
  Auto-GPT &
  \begin{tabular}[c]{@{}l@{}}Content and its \\ vector embeddings\end{tabular} &
  Insert, Select&
  Semi-Structured &
  Non-Symbolic &
  Normal &
  No \\
Matrix-based &
  RMT & 
  \begin{tabular}[c]{@{}l@{}}Memory tokens/\\ memory matrices\end{tabular} &
  Read, Write &
  Semi-Structured &
  Non-Symbolic &
  Low &
  Yes \\ \bottomrule
\end{tabular}}
\end{table}

In this subsection, we provide a comprehensive comparison between ChatDB and recent approaches that augment Transformer-based language models with memory module. 
The memory modules of language models proposed in previous work can be broadly categorized into two types.  
The first type of memory stores context and uses a retrieval model to find content from past interactions that's most relevant to the current conversation, and then uses it as a prompt for the language model~\citep{khattab2022demonstrate}.
We refer to this type of memory as \textit{prompt-based memory}.  
The second type of approach utilizes additional memory tokens or memory matrices as memory~\citep{bulatov2022recurrent}, which we refer to as \textit{matrix-based memory}. 
We compare ChatDB with these approaches based on the following aspects:


\textbf{1. Memory Format.} This aspect relates to the format used for storing memory. ChatDB utilizes databases as its memory.
\textit{Prompt-based memory}~\citep{park2023generative} stores relevant interaction content and/or their corresponding vector embeddings.
\textit{Matrix-based memory} employs additional trainable memory tokens~\citep{bulatov2022recurrent,bulatov2023scaling} or trainable memory matrices~\citep{graves2014neural}.


\textbf{2. Supported Operations.} This aspect refers to the operations supported for manipulating memory. ChatDB supports operations such as insertion, deletion, update, and selection of data in the database memory. The \textit{prompt-based memory} primarily supports insertion and selection operations but lacks complete support for updates and deletions. \textit{Matrix-based memory} supports reading (selection) and writing (insertion, updating, deletion) operations. However, the exact operations performed by the neural network are not explicitly known.


\textbf{3. Memory Storage.} This aspect refers to the format in which data is stored in memory, specifically whether it is structured or not. ChatDB stores memory in a structured format using databases, while both the \textit{prompt-based memory} and \textit{matrix-based memory} are viewed as semi-structured. The reason is that vector embeddings and memory matrices have specific dimensions and sizes, but each individual dimension does not carry a specific and explicit meaning.


\textbf{4. Memory Execution.} This aspect focuses on how memory operations are executed, specifically whether they are symbolic or not. ChatDB executes operations on its database memory using SQL, which is a symbolic language, thus making it inherently symbolic. \textit{Prompt-based memory} performs selections based on a similarity metric using vector embeddings and employs a language encoder to obtain vector embeddings for insertions. Both of these are considered non-symbolic executions. In \textit{matrix-based memory} augmentated LLMs, memory operations are entirely controlled by the neural network, resulting in non-symbolic executions as well.

\textbf{5. Interpretability.} This aspect refers to the degree of interpretability of memory. In ChatDB, memory is stored in a structured and explicit format, and its operations are symbolic, leading to a high level of interpretability. In the \textit{prompt-based memory}, interpretability is generally limited due to inherent challenges in interpreting vector embeddings. For \textit{matrix-based memory} approach, the degree of interpretability is low since memory is completely controlled by the neural network implicitly.


\textbf{6. State Tracking.} This aspect refers to whether the memory effectively tracks the current state of the LLMs. In the case of ChatDB, its memory accurately tracks the current state of the LLMs. The fruit shop experiment serves as a demonstration, where after processing each record, ChatDB's database memory is updated to reflect the latest state of the fruit shop. This showcases how ChatDB's memory effectively tracks its current state. Thanks to the symbolic memory execution, ChatDB's memory allows for easy rollback to any desired timestamp, providing greater flexibility and controllability. In \textit{matrix-based memory} approach, memory is continuously updated and altered by the model itself, enabling it to track the current state of the LLMs. However, the \textit{prompt-based memory} approach simply stores the historical context and only knows what has happened in the past without a clear understanding of the current state.


By examining these aspects, we observe the distinctive features and capabilities of ChatDB in comparison to existing approaches. 
The superiority of ChatDB highlights the advantages of using symbolic memory to augment LLMs.


\section{Evaluation}

In this section, we conduct experiments to evaluate the effectiveness of augmenting LLMs with databases as their symbolic memory. Our experiment results demonstrate that ChatDB significantly outperforms the baseline model ChatGPT, highlighting the advantages of symbolic memory integration.

\subsection{Experimental Setup}
As previously mentioned, using a database as symbolic memory is particularly suitable for scenarios that require precise recording and processing of historical information, such as various data management scenarios. To adapt to the use case of ChatDB and enable quantitative comparisons with other models, we constructed a synthetic dataset simulating the management of a fruit shop.

Furthermore, in order to evaluate the performance of the models, we collected a set of 50 questions with annotated standard answers. These questions varied in difficulty, ranging from hard ones that required multi-hop reasoning to easy ones that only required retrieving information from historical data. There are 15 easy questions and 35 hard questions. Each question is independently answered by the models.

\subsubsection{Model Configuration}
\textbf{ChatDB.}
The LLM used in ChatDB is ChatGPT (GPT-3.5 Turbo), and the hyperparameter \textit{temperature} is set to $0$. We used the MySQL database as the external symbolic memory.

\textbf{Baseline.}
We use ChatGPT (GPT-3.5 Turbo) as the baseline model with a maximum token length of 4096. Similar to ChatDB, we set the \textit{temperature} to $0$.


 \begin{figure*}[p]
    \scriptsize
    \centering
    \begin{tabular}{@{}p{0.475\columnwidth}p{0.475\columnwidth}@{}}
    \toprule
    \textbf{(a) Purchasing goods: the shop purchases goods from suppliers.} 

    \rule[1.2ex]{0.475\columnwidth}{0.4pt}
    
    \textbf{Record:}
    \textcolor[HTML]{3078BE}
    {\textbf{We restocked our store on 2023-01-01 with a new supply of fruits from `ABC' (abc\_sup@example.com, 10080). The purchased quantities include 24 kg cherry, at unit prices of 0.8. Our intended selling price of cherry is 1.3 dollars per unit.}
    }
    
    \textbf{ChatDB:}
    
    Begin to interact with the external database (symbolic memory).

    Step1: Insert supplier `ABC' if not exists
    \begin{minted}[fontsize=\tiny, breaklines]{sql}
INSERT INTO suppliers (supplier_name, contact_number, email)
SELECT 'ABC', '10080', 'abc_sup@example.com'
WHERE NOT EXISTS (SELECT 1 FROM suppliers WHERE supplier_name = 'ABC');
    \end{minted}

    Step2: Insert fruit (set the selling price to NULL and stock quantity to 0) if not exists
    \begin{minted}[fontsize=\tiny, breaklines]{sql}
INSERT INTO fruits (fruit_name, selling_price, stock_quantity, fruit_type, shelf_life)
SELECT 'cherry', NULL, 0, NULL, NULL
WHERE NOT EXISTS (SELECT 1 FROM fruits WHERE fruit_name = 'cherry');
    \end{minted}

    Step3: Insert purchase
    \begin{minted}[fontsize=\tiny, breaklines]{sql}
INSERT INTO purchases (supplier_id, purchase_date, total_cost)
VALUES ((SELECT supplier_id FROM suppliers WHERE supplier_name = 'ABC'), '2023-01-01', 24 * 0.8);
    \end{minted}

    Step4: Insert purchase item
    \begin{minted}[fontsize=\tiny, breaklines]{sql}
INSERT INTO purchase_items (purchase_id, fruit_id, quantity_purchased, cost_per_item, item_total_cost)
VALUES ((SELECT MAX(purchase_id) FROM purchases), (SELECT fruit_id FROM fruits WHERE fruit_name = 'cherry'), 24, 0.8, 24 * 0.8);
    \end{minted}

    Step5: Update the stock quantity of cherry
    \begin{minted}[fontsize=\tiny, breaklines]{sql}
UPDATE fruits 
SET stock_quantity = stock_quantity + 24
WHERE fruit_name = 'cherry';
    \end{minted}

    Step6: Update the selling price of cherry if given new selling price
    \begin{minted}[fontsize=\tiny, breaklines]{sql}
UPDATE fruits 
SET selling_price = 1.3
WHERE fruit_name = 'cherry';
    \end{minted}

    &    
    \textbf{(b) Selling goods: customers buy goods from the shop}

    \rule[1.2ex]{0.475\columnwidth}{0.4pt}
    
    \textbf{Record:}
    \textcolor[HTML]{3078BE}
    {\textbf{A sale was made on 2023-01-02 to `Bob Smith' (contact details: 123-456-7893, bob.smith@example.com). The items purchased were 9 kg apple, 4 kg cherry.}}
    
    \textbf{ChatDB:}
    
    Begin to interact with the external database (symbolic memory).

    Step1: Insert customer 'Bob Smith' if not exists
    \begin{minted}[fontsize=\tiny, breaklines]{sql}
INSERT INTO customers (first_name, last_name, phone_number, email)
SELECT 'Bob', 'Smith', '123-456-7893', 'bob.smith@example.com'
WHERE NOT EXISTS (SELECT 1 FROM customers WHERE phone_number = '123-456-7893');
    \end{minted}

    Step2: Insert sale
    \begin{minted}[fontsize=\tiny, breaklines]{sql}
INSERT INTO sales (customer_id, sale_date, total_price)
VALUES ((SELECT customer_id FROM customers WHERE phone_number = '123-456-7893'), '2023-01-02', (SELECT selling_price FROM fruits WHERE fruit_name = 'apple') * 9 + (SELECT selling_price FROM fruits WHERE fruit_name = 'cherry') * 4);
    \end{minted}

    Step3: Insert sale item
    \begin{minted}[fontsize=\tiny, breaklines]{sql}
INSERT INTO sale_items (sale_id, fruit_id, quantity_sold, price_per_item, item_total_price)
VALUES ((SELECT MAX(sale_id) FROM sales), (SELECT fruit_id FROM fruits WHERE fruit_name = 'apple'), 9, (SELECT selling_price FROM fruits WHERE fruit_name = 'apple'), (SELECT selling_price FROM fruits WHERE fruit_name = 'apple') * 9),
((SELECT MAX(sale_id) FROM sales), (SELECT fruit_id FROM fruits WHERE fruit_name = 'cherry'), 4, (SELECT selling_price FROM fruits WHERE fruit_name = 'cherry'), (SELECT selling_price FROM fruits WHERE fruit_name = 'cherry') * 4);
    \end{minted}

    Step4: Update the stock quantity of apple and cherry
    \begin{minted}[fontsize=\tiny, breaklines]{sql}
UPDATE fruits 
SET stock_quantity = CASE 
    WHEN fruit_name = 'apple' THEN stock_quantity - 9
    WHEN fruit_name = 'cherry' THEN stock_quantity - 4
    ELSE stock_quantity
END
WHERE fruit_name IN ('apple', 'cherry');
    \end{minted}
    
    \\
    \midrule
    \textbf{(c) Return of goods: customers bring back purchased items to the shop.}

    \rule[1.2ex]{0.475\columnwidth}{0.4pt}

    \textbf{Record:}
    \textcolor[HTML]{3078BE}
    {\textbf{Because the customer returned their purchase, we are required to undo the sales transaction made by customer `John Doe' (phone: 123-456-7890, email: john.doe@example.com) on 2023-01-05.}}
    
    \textbf{ChatDB:}
    
    Begin to interact with the external database (symbolic memory).

    Step1: Find the sale\_id for this customer on this date
    \begin{minted}[fontsize=\tiny, breaklines]{sql}
SELECT sale_id FROM sales 
WHERE customer_id = (SELECT customer_id FROM customers WHERE phone_number = '123-456-7890' AND email = 'john.doe@example.com') AND sale_date = '2023-01-05';
    \end{minted}
    \begin{minted}[fontsize=\tiny, breaklines]{python}
Database response:
+---------+
| sale_id |
+---------+
|    9    |
+---------+
    \end{minted}

    Step2: Get all the fruit\_id and quantity\_sold for this sale, replace \textless sale\_id\textgreater \ with the results from the previous queries
    \begin{minted}[fontsize=\tiny, breaklines]{sql}
SELECT fruit_id, quantity_sold FROM sale_items 
WHERE sale_id = 9 [<sale_id>];
    \end{minted}
    \begin{minted}[fontsize=\tiny, breaklines]{python}
Database response:
+----------+---------------+
| fruit_id | quantity_sold |
+----------+---------------+
|    7     |       10      |
|    6     |        9      |
+----------+---------------+
    \end{minted}
    &
    
    Step3: Increase the stock\_quantity for each fruit sold in this sale, replace \textless quantity\_sold\textgreater \  \textless fruit\_id\textgreater \ with the results from the previous queries
    \begin{minted}[fontsize=\tiny, breaklines]{sql}
UPDATE fruits
SET stock_quantity = stock_quantity + 10 [<quantity_sold>]
WHERE fruit_id = 7 [<fruit_id>];

UPDATE fruits
SET stock_quantity = stock_quantity + 9 [<quantity_sold>]
WHERE fruit_id = 6 [<fruit_id>];
    \end{minted}

    Step4: Delete the sale items for this sale, replace \textless sale\_id\textgreater \ with the results from the previous queries
    \begin{minted}[fontsize=\tiny, breaklines]{sql}
DELETE FROM sale_items WHERE sale_id = 9 [<sale_id>];
    \end{minted}

    Step5: Delete the sale record, replace \textless sale\_id\textgreater \ with the results from the previous queries
    \begin{minted}[fontsize=\tiny, breaklines]{sql}
DELETE FROM sales WHERE sale_id = 9 [<sale_id>];
    \end{minted}
    
    \rule[0ex]{0.475\columnwidth}{0.4pt}

    \textbf{(d) Changing prices: modify the price of the goods.}

    \rule[1.2ex]{0.475\columnwidth}{0.4pt}

    \textbf{Record:}
    \textcolor[HTML]{3078BE}
    {\textbf{On 2023-01-05, the sale price of pear in the store was changed to 1.6 dollar per unit.}}
    
    \textbf{ChatDB:}
    
    Begin to interact with the external database (symbolic memory).

    Step1: Update the selling price of pear
    \begin{minted}[fontsize=\tiny, breaklines]{sql}   
UPDATE fruits 
SET selling_price = 1.6
WHERE fruit_name = 'pear';
    \end{minted}

     \\
    \bottomrule
    \end{tabular}
    \caption{Examples showcasing ChatDB's response to the four common operations in the records of the Fruit Shop Dataset. ``9 [\textless sale\_id\textgreater]'' represents that \textless sale\_id\textgreater\ is replaced with 9 in the updated SQL statement. Due to page size limitations, some unnecessary details and ChatDB's summary are omitted.}
    \label{fig:four_manipulations}
    
\end{figure*}


\subsubsection{Dataset}
We synthesize a dataset of fruit shop management records, referred to as the ``Fruit Shop Dataset''. 
This dataset simulates four common operations in a shop: purchasing, selling, changing prices, and goods returns. 
We ensure that all historical records are valid and do not encounter issues such as negative inventory. 
We generate 70 records arranged in chronological order, totaling approximately 3.3k tokens, which is within the maximum token length limit of ChatGPT (4096 tokens).

\textbf{Why do we limit the token length of the dataset?}
If the token length of the dataset exceeds the maximum token length of ChatGPT, memory becomes necessary. 
However, the mainstream memory retrieval methods based on vector embedding are prone to errors. 
This inevitably leads to a decline in the performance of ChatGPT, which is not desired.
Therefore, we deliberately design the token length of the dataset to be within the maximum token length of ChatGPT to avoid using memory and maximize the model's performance. 
Note that ChatDB's performance is generally unaffected by the token length of the dataset. 
Thus, if ChatDB outperforms ChatGPT when the dataset is small, it indicates that ChatDB is also superior to memory-augmented ChatGPT when the dataset is large.

\subsubsection{Processing records}
For ChatDB, the first step is to initialize the database. We need to generate a reasonable database schema for the specific task scenario and create tables in the database. The generation of the database schema can be done manually or using LLMs.
Next, for each record in the dataset, ChatDB processes them one by one. Using the LLM controller, ChatDB manipulates the external database (\textit{i.e.}, symbolic memory) following Algorithm~\ref{alg:chatdb}. We provide examples of ChatDB's response to the four common operations in the Fruit Shop Dataset, namely purchasing, selling, changing prices, and goods returns, as shown in Figure~\ref{fig:four_manipulations}. It is worth emphasizing that ChatDB processes record one by one, so it is not sensitive to the total number of records. Furthermore, each step of the database operation in ChatDB is symbolic and free from errors. Therefore, in theory, ChatDB can handle an infinite number of historical records without sacrificing performance. However, for ChatGPT or existing memory-augmented LLMs, excessively long historical records can significantly degrade performance.
In this experiment, as for the ChatGPT baseline, since the records are not long, we simply treat them as part of the prompt.

\subsubsection{Answering questions}
When answering questions, ChatDB no longer requires records to be part of the prompt. After processing the records, the information is stored in symbolic memory. Following Algorithm~\ref{alg:chatdb}, ChatDB utilizes SQL statements to perform a series of database queries (including calculations) in order to answer the question.
On the other hand, ChatGPT includes records as part of the prompt and directly asks the question. The prompt template is shown in Figure~\ref{fig:eval_prompt_single}.

\begin{figure*}[t]
    \centering
\begin{tabular}{@{}p{\columnwidth}@{}}
\toprule
Suppose you are a fruit shop manager and good at analyzing history records.

The fruit shop newly opened on January 1, 2023. Given the history records for the fruit shop in January 2023, which include customer names, transaction dates, fruit prices, quantities purchased, and whether the items were returned, you need to answer some questions. 

By default, exclude the sales transactions that have been returned when performing calculations.

Here are the historical records of the fruit shop, which are arranged in chronological order based on the occurrence time, surrounded by triple backticks:

\`{}\`{}\`{}

\{\textcolor[HTML]{3078BE}{records}\}

\`{}\`{}\`{}

Based on the history records, answer the question about the fruit shop:

\{\textcolor[HTML]{3078BE}{question}\}
\\
\bottomrule
\end{tabular}
    \caption{Prompt for ChatGPT to answer questions in the Fruit Shop Dataset. The placeholders ``records'' and ``question'' will be replaced by specific details when it is actually used.}
    \label{fig:eval_prompt_single}
\end{figure*}

\subsection{Results}

The experimental results are presented in Table~\ref{tab:main_results}, which clearly demonstrate that ChatDB outperforms ChatGPT with significantly higher accuracy. While ChatGPT is capable of answering easy questions, it falls short in handling hard questions that necessitate multi-hop reasoning and precise calculations. Consequently, ChatGPT exhibits a low accuracy rate for these hard questions. In contrast, ChatDB exhibits a notably high accuracy rate, underscoring the advantages of utilizing a database as symbolic memory. This approach not only prevents error accumulation but also enhances the multi-hop reasoning and precise calculation capabilities of LLMs.

\begin{table}[th]
\centering
\caption{The experimental results of answering questions in the Fruit Shop Dataset. There are 50 questions in total, with 15 being easy questions and 35 being hard questions.
}
\label{tab:main_results}
\begin{tabular}{@{}lrrrr@{}}
\toprule
Model         & Easy & Hard & All & Accuracy \\ \midrule
ChatGPT       &   10/15   &  1/35    &  11/50    &    22\%      \\
ChatDB (ours) &    13/15  &  28/35    &   41/50  &     82\%     \\ \bottomrule
\end{tabular}
\end{table}


 \begin{figure*}[p]
    \scriptsize
    \centering
    \begin{tabular}{@{}p{0.475\columnwidth}p{0.475\columnwidth}@{}}
    \toprule
    \multicolumn{2}{@{}p{\columnwidth}@{}}{
    (a) \textbf{Question:}
    \textcolor[HTML]{3078BE}{\textbf{What was the total revenue for January 2023?}}}\\
    \midrule
    \textbf{ChatGPT:}
    
To calculate the total revenue for January 2023, we need to sum up the revenue generated from each sales transaction. We can use the following formula to calculate the revenue for each transaction: revenue = quantity * selling price

We also need to exclude the sales transactions that have been returned. Based on the history records, the following sales transactions have been returned:

- Alice Smith on 2023-01-17 - Bob Smith on 2023-01-05 \textcolor[HTML]{808080}{[...]}





Therefore, we need to exclude these transactions when calculating the total revenue.
Here is the calculation for each sales transaction:

- Bob Smith on 2023-01-02: (9 kg * 3.8 USD/kg) + (4 kg * 1.3 USD/kg) = 39.1 USD \textcolor{red}{[WRONG]: calculation error, should be 39.4}

- Sue Davis on 2023-01-02: 1 kg * 3.8 USD/kg = 3.8 USD

- Bob Smith on 2023-01-05: (5 kg * 1.8 USD/kg) + (2 kg * 3.0 USD/kg) + (6 kg * 2.8 USD/kg) = 47.6 USD \textcolor{red}{[WRONG]: calculation error and this transaction has been returned.}


    \textcolor[HTML]{808080}{[...]}  \textcolor{red}{[WRONG]: most of the sale transactions are calculated incorrectly.}
    
    \textcolor{red}{[STOP]: Because it exceeds the maximum token length.}
    &    
    \textbf{ChatDB:}

    Begin to interact with the external database (symbolic memory).

    Step1: Calculate the total revenue for January 2023
    
    \begin{minted}[fontsize=\tiny, breaklines]{sql}   
SELECT SUM(total_price) AS total_revenue
FROM sales
WHERE sale_date >= '2023-01-01' AND sale_date < '2023-02-01';
    \end{minted}
    \begin{minted}[fontsize=\tiny, breaklines]{python}
Database response:
+---------------+
| total_revenue |
+---------------+
|     707.0     |
+---------------+
    \end{minted}
    \\
    \midrule
    \multicolumn{2}{@{}p{\columnwidth}@{}}{
    (b) \textbf{Question:}
    \textcolor[HTML]{3078BE}{\textbf{Which day in January 2023 had the highest revenue?}}}\\
    \midrule
    \textbf{ChatGPT:}

    To calculate the revenue for each day, we need to sum up the total sales for that day (excluding any returned items) and multiply by the selling price. We can then compare the revenue for each day to determine which day had the highest revenue.

Here is the breakdown of sales for each day:

~

- January 2: 9 kg apple at 3.8 = 34.2, 4 kg cherry at 1.3 = 5.2, 3 kg cherry at 1.3 = 3.9, 3 kg orange at 2.1 = 6.3, total sales = 49.6 \textcolor{red}{[WRONG]: should be 43.2}

- January 4: 2 kg apple at 1.5 = 3, 4 kg pear at 1.6 = 6.4, 7 kg cherry at 1.3 = 9.1, total sales = 18.5 \textcolor{red}{[WRONG]: should be 27.9}

- January 5: 5 kg cherry at 1.9 = 9.5, 10 kg orange at 1.8 = 18, 1 kg cherry at 1.9 = 1.9, 1 kg grape at 3.2 = 3.2, total sales = 32.6 \textcolor{red}{[WRONG]: should be 83}

- January 6: 3 kg apple at 1.5 = 4.5, 10 kg grape at 3.2 = 32, total sales = 36.5 \textcolor{red}{[WRONG]: should be 29.4}

- January 7: 4 kg pear at 1.6 = 6.4, total sales = 6.4 \textcolor{red}{[WRONG]: should be 37.2}

- January 8: 6 kg peach at 3 = 18, total sales = 18  \textcolor{red}{[WRONG]: should be 21.6}







~

\textcolor[HTML]{808080}{[...]}  \textcolor{red}{[WRONG]: most of the daily revenues are calculated incorrectly.}

\textcolor{red}{[STOP]: Because it exceeds the maximum token length.}

    &
    \textbf{ChatDB:}
    
    Begin to interact with the external database (symbolic memory).

    Step1: Calculate the total revenue for each day in January 2023
    
    \begin{minted}[fontsize=\tiny, breaklines]{sql}   
SELECT sale_date, SUM(total_price) AS revenue
FROM sales
WHERE sale_date BETWEEN '2023-01-01' AND '2023-01-31'
GROUP BY sale_date;
    \end{minted}
    \begin{minted}[fontsize=\tiny, breaklines]{python}
Database response:
+------------+---------+
| sale_date  | revenue |
+------------+---------+
| 2023-01-02 |   43.2  |
| 2023-01-04 |   27.9  |
|     ...    |    ...  |
+----------------------+
    \end{minted}
    
    Step2: Find the day with the highest revenue
    
    \begin{minted}[fontsize=\tiny, breaklines]{sql}   
SELECT sale_date, SUM(total_price) AS revenue
FROM sales
WHERE sale_date BETWEEN '2023-01-01' AND '2023-01-31'
GROUP BY sale_date ORDER BY revenue DESC LIMIT 1;
    \end{minted}
    \begin{minted}[fontsize=\tiny, breaklines]{python}
Database response:
+------------+---------+
| sale_date  | revenue |
+------------+---------+
| 2023-01-30 |  119.7  |
+------------+---------+
    \end{minted}
    \\
    \midrule
    \multicolumn{2}{@{}p{\columnwidth}@{}}{
    (c) \textbf{Question:}
    \textcolor[HTML]{3078BE}{\textbf{What was the average weight of fruit per sales transaction in January 2023?}}}\\
    \midrule
    \textbf{ChatGPT:}
To calculate the average weight of fruit per sales transaction, we need to first identify all the sales transactions in January 2023 and calculate the total weight of fruit sold in those transactions. Then, we can divide the total weight by the number of sales transactions to get the average weight per transaction.

Here are the sales transactions in January 2023:

- Bob Smith on 2023-01-02: 9 kg apple, 4 kg cherry

- Sue Davis on 2023-01-02: 1 kg apple

- Bob Smith on 2023-01-05: 5 kg orange, 2 kg peach, 6 kg pear \textcolor{red}{[WRONG]: this transaction has been returned.}

- Sue Davis on 2023-01-05: 5 kg pear, 10 kg orange, 1 kg cherry

- Alice Smith on 2023-01-05: 5 kg cherry

\textcolor[HTML]{808080}{[...]}  \textcolor{red}{[WRONG]: not find out all valid sale transactions.}

To calculate the total weight of fruit sold, we can add up the quantities of each fruit sold in the above transactions:

- Apples: 16 kg
- Cherries: 12 kg
- Oranges: 10 kg
- Pears: 18 kg
- Kiwis: 12 kg
- Grapes: 18 kg
- Bananas: 27 kg
- Strawberries: 7 kg
- Peaches: 9 kg
- Plums: 8 kg

The total weight of fruit sold in January 2023 is 127 kg. \textcolor{red}{[WRONG]: It should be 278. Even according to the calculation above, the total should be 137, not 127.}

To calculate the average weight per sales transaction, we need to divide the total weight by the number of sales transactions. There are 18 sales transactions in January 2023, \textcolor{red}{[WRONG]: It should be 28 valid sale transactions in total.} so the average weight \textcolor{red}{[STOP]: Because it exceeds the maximum token length.}
    
    &
    \textbf{ChatDB:}
    
    Begin to interact with the external database (symbolic memory).

    Step1: Calculate the total weight of fruit sold in January 2023
    
    \begin{minipage}{0.32\columnwidth}
        \begin{minted}[fontsize=\tiny, breaklines]{sql}   
SELECT SUM(si.quantity_sold) as total_weight
FROM sale_items si
JOIN sales s ON si.sale_id = s.sale_id
WHERE s.sale_date BETWEEN '2023-01-01' AND '2023-01-31';
        \end{minted} 
    \end{minipage}
    \begin{minipage}{0.14\columnwidth}
        \begin{minted}[fontsize=\tiny, breaklines]{python}
Database response:
+--------------+
| total_weight |
+--------------+
|     278      |
+--------------+
        \end{minted}
    \end{minipage}

    \vspace{0.1cm}
    Step2: Count the number of sales transactions in January 2023
    
    \begin{minipage}{0.32\columnwidth}
        \begin{minted}[fontsize=\tiny, breaklines]{sql}   
SELECT COUNT(DISTINCT s.sale_id) as num_sales
FROM sales s
WHERE s.sale_date BETWEEN '2023-01-01' AND '2023-01-31';
        \end{minted}
    \end{minipage}
    \begin{minipage}{0.14\columnwidth}
        \begin{minted}[fontsize=\tiny, breaklines]{python}
Database response:
+-----------+
| num_sales |
+-----------+
|     28    |
+-----------+
        \end{minted}
    \end{minipage}

    Step3: Calculate the average weight of fruit per sales transaction in January 2023, replace \textless total\_weight\textgreater\ and \textless num\_sales\textgreater\ with the results from the previous queries

    \begin{minipage}{0.29\columnwidth}
        \begin{minted}[fontsize=\tiny, breaklines]{sql}   
SELECT 278 [<total_weight>] / 28 [<num_sales>] 
as avg_weight_per_sale;
        \end{minted}
    \end{minipage}
    \begin{minipage}{0.18\columnwidth}
        \begin{minted}[fontsize=\tiny, breaklines]{python}
Database response:
+---------------------+
| avg_weight_per_sale |
+---------------------+
|         9.93        |
+---------------------+
        \end{minted}
    \end{minipage}
    
    \\
    \bottomrule
    \end{tabular}
    \caption{
    Examples of ChatDB and ChatGPT answering questions. While ChatGPT fails to answer any of the questions correctly, ChatDB successfully answers all of them. The issues with ChatGPT are highlighted in red. Some irrelevant details and ChatDB's summary are omitted for brevity. The gray ellipsis \textcolor[HTML]{808080}{[...]} indicates that the response is trimmed.
    }
    \label{fig:question_examples}
    
\end{figure*}


We present several examples of the two models answering questions in Figure~\ref{fig:question_examples} for comparison. In all these examples, ChatDB correctly answers the questions while ChatGPT fails. ChatGPT often exhibits errors in calculating the total price of each sale transaction, as observed in Figure~\ref{fig:question_examples}(a).
Sometimes, the formulas are correct but the calculations are wrong, while other times, even the formulas are incorrect.
In addition, ChatGPT struggles to find all valid sale transactions, leading to errors in its answering process. This issue is commonly encountered and evident in all these examples. Furthermore, ChatGPT tends to make sequential errors, resulting in significant error accumulation.

In contrast, ChatDB performs quite well in these examples. During the initial processing of records, symbolic operations (\textit{i.e.}, SQL operations) are applied to manipulate the database (\textit{i.e.}, symbolic memory), ensuring that all information is stored in a structured form within the database. When answering questions, ChatDB generates SQL statements to query the database. The three examples demonstrate the effectiveness of ChatDB in solving problems requiring one, two, and three chain-of-memory steps, respectively. We can observe that ChatDB accurately answers the questions, and the execution logic of the chain-of-memory is clear, with each step tightly connected and approaching the ultimate answer.
From these examples, the advantages of ChatDB are evident in two aspects:

1. Through the chain-of-memory approach, complex problems are decomposed into multiple steps of memory operations, simplifying the problem's complexity. Each step's result is accurately stored as an intermediate outcome and used in subsequent steps, which greatly assists in complex reasoning.

2. Symbolic memory enables precise operations and calculations. ChatDB delegates many calculational tasks to the external database by executing SQL statements, ensuring the accuracy of each step and preventing error accumulation.

In summary, by leveraging external databases as symbolic memory, ChatDB significantly outperforms ChatGPT in this experiment.


\section{Conclusion}
In this paper, we introduce ChatDB, a framework that augments LLMs with symbolic memory in the form of databases.
We demonstrate the advantages and capabilities of symbolic memory and chain-of-memory approach in enhancing complex reasoning and preventing error accumulation. 
By providing a precise storage mechanism for intermediate results, symbolic memory enables accurate and reliable operations. 
Moreover, the use of symbolic languages, such as SQL, allows symbolic computation and manipulation of stored information. 
Through experimental evaluation, we observe a significant improvement in performance with ChatDB compared to ChatGPT. 
The integration of symbolic memory in ChatDB substantially enhances the model's ability to handle various queries and reasoning tasks in management settings. 
This improvement highlights the benefits and effectiveness of leveraging symbolic memory in LLMs.

\bibliography{chain_memory}
\bibliographystyle{apalike}

\end{document}